\documentclass[journal]{IEEEtran}
\ifCLASSINFOpdf
\else
\fi
\usepackage[dvips]{graphicx}

\begin{document}
%
\title{HAR-Net: Joint Learning of Hybrid Attention for Single-stage Object Detection}
%
%
%

\author{Ya-Li~Li,~\IEEEmembership{Member,~IEEE,}
        and~Shengjin~Wang,~\IEEEmembership{Member,~IEEE}
\thanks{Ya-Li Li and Shengjin Wang are with the Department
of Electrical Engineering, Tsinghua University, Beijing, 100084 PRC. e-mail: liyali13, wgsgj@tsinghua.edu.cn.}
\thanks{This work was supported by the National Natural Science Foundation of China under Grant Nos. 61701277, 61771288 and the state key development program in 13th Five-Year under Grant No. 2016YFB0801301.}
\thanks{Manuscript received April 12, 2019; revised August 26, 2019.}}

%
%

\markboth{Journal of \LaTeX\ Class Files,~Vol.~14, No.~8, April~2019}%
{Shell \MakeLowercase{\textit{et al.}}: Bare Demo of IEEEtran.cls for IEEE Journals}
%



\maketitle

\begin{abstract}
Object detection has been a challenging task in computer vision. Although significant progress has been made in object detection with deep neural networks, the attention mechanism is far from development. In this paper, we propose the hybrid attention mechanism for single-stage object detection. First, we present the modules of spatial attention, channel attention and aligned attention for single-stage object detection. In particular, stacked dilated convolution layers with symmetrically fixed rates are constructed to learn spatial attention. The channel attention is proposed with the cross-level group normalization and squeeze-and-excitation module. Aligned attention is constructed with organized deformable filters. Second, the three kinds of attention are unified to construct the hybrid attention mechanism. We then embed the hybrid attention into Retina-Net and propose the efficient single-stage HAR-Net for object detection. The attention modules and the proposed HAR-Net are evaluated on the COCO detection dataset. Experiments demonstrate that hybrid attention can significantly improve the detection accuracy and the HAR-Net can achieve the state-of-the-art 45.8\% mAP, outperform existing single-stage object detectors. 
\end{abstract}


%
\IEEEpeerreviewmaketitle

\section{Introduction}

\IEEEPARstart{O}{bject} detection is a fundamental and challenging issue in computer vision. It is an important part of visual understanding. In recent years, we have witnessed that Convolutional Neural Networks (CNN)~\cite{LeCun_cnn89} have brought tremendous progress to the field of object detection. Influential methods include Faster-RCNN~\cite{Ren_fasterrcnn15}, R-FCN~\cite{Dai_rfcn16}, YOLO~\cite{Redmon_yolo16}\cite{Redmon_yolo9000_17}, SSD~\cite{Liu_ssd14}, light-head RCNN~\cite{Li_lightrcnn17}, FPN~\cite{Lin_FPN17}, Retina-Net~\cite{Lin_retina17}. More specificly, highly efficient single-stage object detection methods with fully-convolutional networks are proposed. Convolution layers with non-linear activations are stacked into a feedforward architecture for feature representative learning. Spatial pooling is embedded to enlarge the receptive field. Early layers of CNN extract fine details such as edges, lines and corners. Last layers obtain coarse semantic representations. Due to the varying size of objects, multi-level features are usually fused by top-down connections for object detection.

\begin{figure}[htbp]
\centering
\includegraphics[width=3.6in]{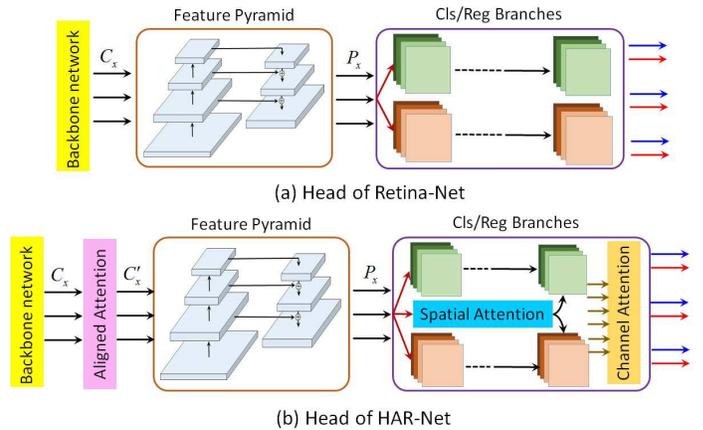}
\caption{Overview of hybrid attention embedded single-stage object detection. Three components constitute into the hybrid attention. a) \textbf{Aligned attention} is used for cross-level feature alignment in pyramid construction. b) \textbf{Channel attention} is plugged for multi-scale feature re-weighting and selection. c) \textbf{Spatial attention} is applied to softly localize interested regions and parts for detection.
{\label{fig_harnet}}}
\end{figure}

Generally, CNN-based object detection methods can be divided into two kinds. The first kind finds objects of different categories in a sliding-window way. Input images are gridded and the anchors correspond to spatially-localized regions are assigned. CNN outputs of those regions are used to discriminate objects from backgrounds. These so-called single-stage detectors are simple and efficient in design and training, yet suffer from serious imbalanced classification issue. The second kind builds region proposal network (RPN) as one branch to find category-independent candidate regions of interest (ROIs). Then the ROI-pooling layer is added to generate fixed-length feature representations. Another branch of network is learned to classify and localize the objects of various classes. These multi-stage detectors always need additional ROI-Pooling operations for better localization performance.

In this paper, we aim to incorporate visual attention with single-stage object detection. Compared with RPN-based methods, the single-stage detectors significantly suffers from imbalanced classification and imprecise feature representation issues. Although various techniques such as OHEM~\cite{Shrivastava_ohem16}, focal loss~\cite{Lin_retina17} and gradient harmonized mechanism~\cite{Li_ghm19} are proposed to attend hard examples for effective training, the attention mechanism in network design is still far from exploitation. Our work is different from existing methods since we embed attention modules into the network architecture. The overview of the network is illustrated in Fig.~\ref{fig_harnet}. The visual attention is formulated as a sequence of various attention modules to softly re-organize and re-weight the feature response maps. First, spatial attention with stacked dilated convolutions is proposed to re-weight feature response maps inside larger receptive fields. It can be viewed as soft embedded region proposal in single-stage detection. Second, channel attention is learned based on cross-level group normalization (CLGN) and global feature based squeeze-and-excitation. It performs feature selection by attached re-scaling. Third, aligned attention is imposed as an independent module with deformable convolution filters, for pyramidal feature alignment. We sequentially combine the three kinds of attention into the hybrid attention for single-stage object detection. With the hybrid attention, the object detection accuracy can be improved with limited extra storage and parameters.

To summarize, the contributions of this work are as follows:

(1) We fully investigate different kinds of attention and propose a hybrid attention mechanism for single-stage object detection. In particular, the hybrid attention is formulated as the sequential combination of spatial attention, channel attention and aligned attention.

(2) We propose the HAR-Net to effectively integrate the hybrid attention into Retina-Net. Weight sharing of attention modules in pyramidal paradigm is studied. Multi-scale testing is presented to further improve the detection performance.

(3) Experiments using different backbone networks show that the detection performance can be significantly improved with the hybrid attention. We also achieve the state-of-the-art single-stage detection accuracy on COCO \textit{test-dev} dataset.

The remainder part of this paper is organized as follows. In Section~\ref{sec:rel}, we briefly introduce the related work, such as object detection and attention networks. The proposed hybrid attention modules are presented in Section~\ref{sec:att}. In Section~\ref{sec:net}, we present the details of network architecture with embedded hybrid attention. Experimental results for validation are given in Section~\ref{sec:exp}. Section~\ref{sec:con} concludes the paper.


\section{Related work}\label{sec:rel}
CNNs~\cite{LeCun_cnn89} have been proven effective in tackling a variety of visual tasks, including image classification~\cite{Krizhevsky_imagenet_cnn12}\cite{He_resnet16}, object detection~\cite{Girshick_rcnn14}\cite{Girshick_fastrcnn15}\cite{Lin_FPN17}\cite{Redmon_yolo16}\cite{Redmon_yolo9000_17}, semantic and instance segmentation~\cite{Long_fcn15}\cite{He_maskrcnn17}. Here we present a brief review on the object detection methods and the attention mechanism in CNNs.

\subsection{Object detection}
In recent decades, research on object detection has been prospering, mainly due to the advances in deep learning and the practical importance. From the feature perspective, the object detection methods can be divided into the classical hand-crafted-feature-based detection and CNN-based detection. The CNN-based object detection can be further divided into as single-stage detectors with sliding-window paradigm and multi-stage detectors with region proposals.

\textbf{Hand-crafted-feature-based object detection.} Deformable Part Model (DPM)~\cite{Felz10} is one of the most influential object detection methods with hand-crafted features. It represents the objects with multiple parts. The object appearances are represented by multi-level HOG (\textit{i.e.}, histograms of gradients) features and the deformations of parts are formulated into latent variables. Latent support vector machine (SVM) is employed to learn the parameters of this mixture model. Besides of HOG, a variety of hand-designed features are applied into object detection. For example, Boosted-LBP and HOG are combined for detection~\cite{Zhang10}. In~\cite{Khan13}, color names into HOG features are fused. Generic mid-level part dictionary for feature sharing~\cite{Pirsiavash12}\cite{Girshick13} are learned to improve the detection accuracy of the traditional methods.

\textbf{Region-proposal based object detection.} For these object detectors, at least two stages are essential for effective learning. In the first stage, a small number of candidate region proposals are generated with relatively low computation cost. These candidates are supposed to contain interested category-independent objects. In the second stage, classifiers are learned and performed on the candidate regions to identify the categories of objects. Multiples stages can be cascaded to refine the detections~\cite{Cai_cascadeRCNN18}. Traditional methods for region proposals include Selective Search~\cite{Uijlings13}, BING~\cite{Cheng_bing14}, EdgeBoxes~\cite{Zitnick14}. Moreover, RPN is proposed and integrated into Faster-RCNN~\cite{Ren_fasterrcnn15} to learn the end-to-end network for detection. RPN takes an image of arbitrary size as input and outputs a set of region proposals with objectness scores. Another branch of network is added to extract translation-invariant per-region features. The classification and bounding box regression outputs are obtained with the per-region features to predict object categories and refine the locations. In Faster-RCNN~\cite{Ren_fasterrcnn15}, a network branch named Fast-RCNN with ROI-pooling operation and stacked fully-connected layers is trained for object category and location prediction. The RPN and Fast-RCNN branches share the backbone network for feature extraction in order to save the computational cost. R-FCN~\cite{Dai_rfcn16} is another two-stage object detector which applies the position-sensitive ROI-pooling to tackle the dilemma between translation-invariance in classification and translation-variance in localization. Light-head R-CNN~\cite{Li_lightrcnn17} applies ROI warping in thin feature maps generated by separable convolutions for larger receptive fields with few parameters. Multi-path network~\cite{Zagoruyko_multipath16} and cascade R-CNN~\cite{Cai_cascadeRCNN18} presents parallel and cascade stages of network branches for improving the classification and localization accuracy.

\textbf{Box generation-based object detection.} This kind of object detectors directly predicts the class probabilities and the coordinates of bounding boxes. Without proposal generation, images are gridded or anchor boxes are attached with indexes in feature maps to localize the objects. In YOLO~\cite{Redmon_yolo16}, images are gridded into several (\textit{i.e.}, $7 \times 7$) regions. The probabilities of multiple categories are predicted for each divided region and CNN outputs from several regions are combined for final object localization. The highly-efficient YOLO is extended to detect objects from a large range of categories (over 10,000) with higher accuracy in small objects, named YOLO 9000~\cite{Redmon_yolo9000_17} and YOLO-v3~\cite{Redmon_yolov3_18}, respectively. SSD~\cite{Liu_ssd14} spreads out the boxes from different scales into multiple layers and combines the predictions from multi-level feature maps to handle objects of various sizes. DSSD~\cite{Fu_DSSD17} adds deconvolution module to increase the resolution of feature maps for detection refinement. Retina-net~\cite{Lin_retina17} predicts object existence over densely sampled regions, by anchors correspond to boxes with different sizes and aspect ratios. It applies the focal loss to prevent the vast number of easy negatives from overwhelming the detector in training. Besides of focal loss, gradient harmonized mechanism (GHM)~\cite{Li_ghm19} is introduced to reduce the impacts of easy negatives for effect learning. Anchor-free module~\cite{Zhu_fsaf19} is proposed for box encoding and decoding at arbitrary level in feature pyramid, as well as online feature selection to improve the single-stage accuracy.

By comparison, the multi-stage detectors apply RPN to eliminate an enormous number of candidates to alleviate the foreground/background imbalance. The ROI-Pooling layers are then added to extract scale-translation invariant features for effective classification and regression. Thus they are complex in design with low efficiency. In the other way, single-stage detectors are simpler in design and training, leading to high efficiency. The gridding-based YOLO and SSD are quite fast detectors, but the detection accuracy is relatively lower than multi-stage detectors. The serious class imbalance issue and imprecise feature representations are the main obstacles to impede the detection accuracy. In this work, we motivate to tackle the two main issues with attention mechanism and promote the single-stage detection.

\subsection{Attention for visual recognition}
The natural basis of attention mechanism has been extensively studied, due to its importance in perception and cognition. In general, attention aims to select the focused locations or important feature representations at fixed locations. Visual attention has been incorporated with feedforward network architecture in an end-to-end training way for image classification. For example, Residual Attention Network~\cite{Wang_ran17} consists of residual modules and attention modules to generate attention-aware feature representations. The ¡°Squeeze-and-Excitation¡±(SE) blocks in SE-Nets~\cite{Hu_senet18} can be viewed as the channel attention, which assign learned weights for different channels of convolutional layers and recalibrates feature response maps adaptively in the channel-wise way, to improve visual recognition performance.

Attention is widely researched for visual understanding tasks, such as visual question answering (VQA) and image captioning. Salient region detection is used to fix the gaze for image captioning~\cite{Xu_showattendtell16}. In~\cite{Xu_ask2016}, a spatial attention architecture to align words with image patches and another attention to consider the semantic question as well as choose visual evidence is combined for VQA. The attention mechanism is embedded into the recurrent neural networks (RNN), to connect encoder and decoder for sequence transduction~\cite{Vaswani_attendisallneed17}. 	SCA-CNN~\cite{Chen_scaCNN18} combines multi-layer spatial-channel attention with different orders for effective encoder for image captioning. Multi-modal attention such as image and question attention is jointly learned with a bilinear pooling structure for VQA~\cite{Yu_MFB2017}. In~\cite{Anderson_bottomup18}, object detection proposes image regions and serves as the bottom-up attention. It is combined with top-down attention which assigns weights for collected regions for image captioning and VQA. Generally, the attention for visual understanding tasks mainly focuses on the decoder phase with RNN architecture.

There are efforts to introduce attention into object detection. The RPN in multi-stage object detectors~\cite{Ren_fasterrcnn15}\cite{Dai_rfcn16}\cite{Li_lightrcnn17} which selects regions of interests can be viewed as the ¡°hard¡± spatial attention. In AttractioNet~\cite{Gidaris_attractionNet16}, an active box proposal generation strategy is proposed to progressively focus on the promising image areas for region proposals, behaving like an attention mechanism. These works are hard spatial attention for box selection. Although RPN is effective for find the candidates, the hard selection of possible regions would cause missing detections, leading to the decrease of recall rate. Besides, the object detection network with an additional RPN is heavy in both parameter size and time cost.

In this paper, we focus on the attention mechanism for single-stage object detection, especially the simple attention modules which can be easily plugged into CNNs. We sequentially integrate different kinds of soft attention into CNNs for effective detection. The spatial attention learns to reweight the regions with box-wise objectness scores and the channel attention learns channel-wise weights for soft feature selection. The aligned attention uses deformation to refine the feature representations. These attention modules are fused into the hybrid attention to improve the single-stage object detection performance.

\section{Hybrid attention}\label{sec:att}

\begin{figure}[htbp]
\centering
\includegraphics[width=3.0in]{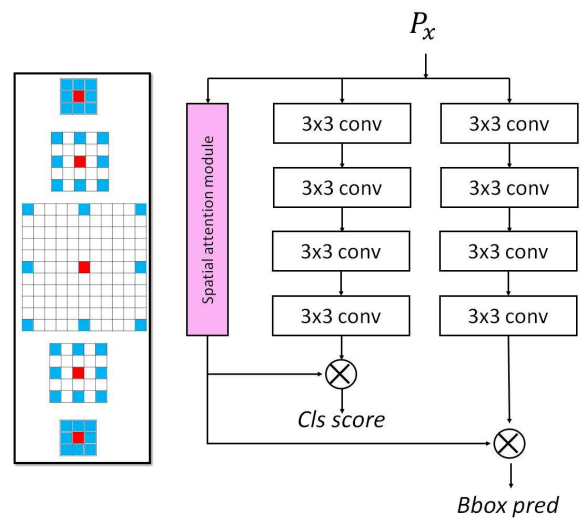}
\caption{Illustration of stacked dilated convolution based spatial attention. The dilation rates of convolutional layers is firstly increased to enlarge the receptive field, then decreased to reduce the gridding artifacts. This design avoids the misalignment of sequential downsampling and upsampling.{\label{fig_spa}}}
\end{figure}

\begin{figure*}[htbp]
\centering
\includegraphics[width=5.4in]{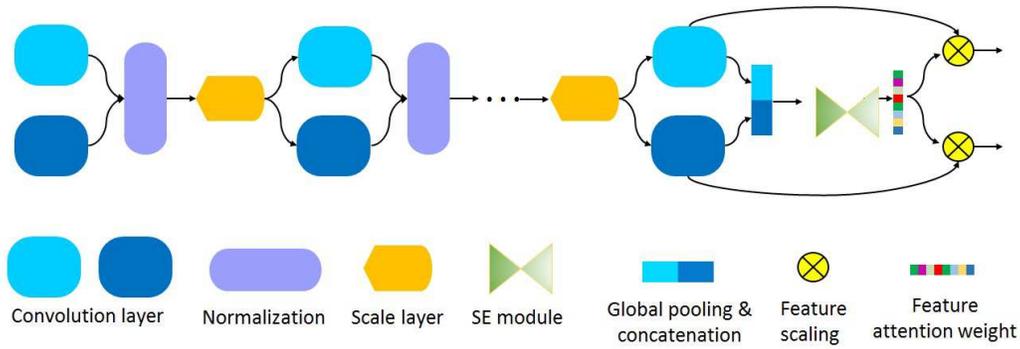}
\caption{Pyramid-sharing channel attention. The channel attention of middle convolutional layers is modeled with $L_2$ group normalization and attached feature scaling. We add a squeeze-excitation (SE) block after the last convolution to learn the channel importance with global information. In pyramid structure, the feature scaling parameters are shared across multi-level feature representations to promote the information flow. {\label{fig_fea}}}
\end{figure*}

Deep CNNs output hierarchical features by extracting and fusing spatial and channel-wise information in a unified way. Single-stage detectors use CNN features to directly classify whether a specified region contain object or not. We introduce multiple visual attention modules into single-stage object detection for three reasons. First, the classification in single-stage object detection is highly imbalanced. It needs to discriminate objects from numerous complex backgrounds. Thus attending important regions and features can alleviate the imbalance. Second, visual attention can impose global context information to select and produce more effective features for detection. Third, attention helps use the features of focus for attended regions to update the network parameters.

Based on the above considerations, we propose and integrate the hybrid attention for single-stage object detection. Our attention mechanism mainly focuses into three types, such as \textit{spatial attention}, \textit{channel attention} and \textit{aligned attention}. \textit{Spatial attention} is designed by stacked dilated convolutional layers to generate a soft mask for weighting multi-level features. \textit{Channel attention} is implemented by adaptive feature scaling with the cross-level global and local features. \textit{Aligned attention} is designed into an independent deformable module for the alignment of high-level coarse features. Since feature pyramid with bottom-up and top-down connections is essential for high-performance multi-scale object detection~\cite{Liu_ssd14}~\cite{Lin_FPN17}~\cite{Lin_retina17}, we further set the spatial attention to be shared across multiple tasks and channel attention to be shared across multiple pyramid levels. We refer them as \textit{task-sharing spatial attention} and \textit{pyramid-sharing channel attention}, respectively.

\subsection{Spatial attention}

Spatial attention for object detection can be viewed as control gating to filter less important background regions. For multi-stage object detectors, RPN can be considered as the hard spatial attention to pre-select regions, but it might cause missing detections. For single-stage detection, we design the soft spatial attention to weight points of features with learned probability masks. There are works to apply bottom-up top-down structure with sequential downsampling/upsampling for spatial attention masks in image classification~\cite{Wang_ran17}. Downsampling with max-pooling rapidly gathers the global context and upsampling generates the masks with equal size for easy combination. However, it is difficult to apply in single-stage detection. Max-pooling would cause misalignment artifacts for feature maps whose size is indivisible. This misalignment issue is hard to avoid since the image sizes vary a lot. To conquer this, we propose to use the stacked dilated convolutions for spatial attention learning.

Our spatial attention module is constructed by a symmetrical organization of stacked dilated convolutions, as illustrated in Fig.~\ref{fig_spa}. Convolutional layers with different dilation rates are sequentially stacked. At the beginning, the dilation rates of convolution layers are increased to enlarge the receptive field and collect the global information. Then the dilation rates are progressively decreased to estimate the relative importance of spatial feature maps within the global context region, as well as reduce the gridding mosaic artifacts. The last convolutional layer outputs a 1-channel mask with attention scores. With \textit{sigmoid} activation, the module generates a feature map with each pixel indicating the relative spatial importance. We follow the hybrid dilated convolution rule in the work of semantic segmentation~\cite{Yu_DRN17}~\cite{Wang_hdc18}, to further avoid gridding artifacts. To simply the design, the channels of the dilated convolution is set the same except for the last convolution layer, usually $C=16$.

The outputs of the spatial attention module are the soft spatial weights, which are further multiplied with the original feature values. Suppose the output of spatial attention module is $A^s (i,j)$, and the feature response maps are $F(c,i,j)$, then the spatially weighted features are denoted as
\begin{equation}
A^s (i,j) * F(c,i,j)
\label{eq1}
\end{equation}
where $i,j$ and $c$ are spatial and channel indexes, respectively. For object detection, two branches for classification and bounding box regression are learned in parallel. We share the learned spatial attention in this multi-task learning, as \textit{task-sharing spatial attention}. In the feedforward way, the spatial attention can degrade the information from backgrounds therefore alleviate the imbalanced classification of single-stage detection. In the feedbackward way, the useful features for object detection can be learned in an enhanced way with the focused gradient updating. Besides, we add the spatial attention module as an additional network branch in parallel with the feature learning network. It would widen the network and further improve the feature representative learning for object detection. By taking the global information into account, the spatial importance can be well estimated with this module.

\subsection{Channel attention}\label{sec:fe_att}

Channel attention can be viewed as the selection of feature response maps for high-level abstraction on the demand of object detection. We design the channel attention to assign soft weights for convolutional feature maps, in which the weights indicate the channel-wise feature importance. Our proposed channel attention module is shown in Fig.~\ref{fig_fea}. First, we impose $L_2$ group normalization after each convolution layer. For pyramidal object detection with sharing convolutional parameters, we introduce the cross-level group normalization (CLGN). That is, the normalization imposed for groups of channels across the pyramid to estimate the relative importance. Feature scaling is applied after the CLGN for channel re-weighting. For the last non-decision convolutional layer, we use the squeeze-and-excitation (SE) block to extract the channel relations with the global information. The global features across different channels are firstly aggregated through the squeeze operation. Then the activations for each channel by the mutual dependence are learned by the excitation operation. For the pyramid, we concatenate the global information from different streams together to learn the spatial attention. That is, we introduce cross-level squeeze-and-excitation (CLSE) for \textit{pyramid-sharing channel attention}. The global max-pooling is firstly used to obtain a vector with the length equal to channels number. The output vectors from different levels are then concatenated together into a long vector and the SE-block is used to learn activated weights for different channels with the concatenated vector. The learned weights indicate the channel activation importance, thus corresponds to the channel attention weights for further feature scaling.

In particular, we share the channel attention across different pyramid levels to utilize the multi-scale global context, but separate it for different tasks in detection. That is, we learn the channel attention for the classification and bounding box regression branches, respectively. Suppose the channel attention is $A^{cls/bbox}(c)$, we directly multiply it with the output feature maps. The channel attended feature maps are with the values as
\begin{equation}
A^{cls/bbox}(c)*F(c,i,j)
\label{eq2}
\end{equation}
Unlike spatial attention, channel attention performs the feature selection for effective learning. In the bottom-up way, channel attention selects important attributes for high-level feature learning on the demand of detection, meanwhile suppresses less informational channels. In the top-down way, the feature maps are selectively updated with the object guidance. Moreover, existing research has demonstrated that by stacking the SE-blocks together, the generalization performance of deep residual CNNs for visual recognition can be improved~\cite{Hu_senet18}. We employ the ImageNet pre-trained SE-Nets as the backbone networks to combine channel attention into backbone feature representation.

\subsection{Aligned attention}

\begin{figure}[htbp]
\centering
\includegraphics[width=3.4in]{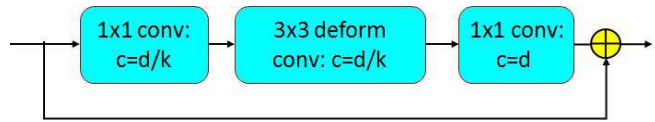}
\caption{Aligned attention module. The module is constructed by $3\times 3$ deformable convolution for feature alignment, with attached $1\times 1$ convolutions for channel organization.  {\label{fig_align_att}}}
\end{figure}

Aside from spatial and channel attention, we introduce the aligned attention in feature pyramid construction, for precise feature representation. In feature pyramid, multi-level features from the backbone network are firstly generated by bottom-up connections then fused by top-down lateral connections. High-level features with lower resolutions are upsampled and merged with low-level ones with higher resolution, for the purpose of detecting small objects with contexts. The aligned attention aims to model the transformations of features from various levels for flexible organization. We use the deformable convolution to model the aligned attention. Existing work~\cite{Dai_deform17} replaces the $3 \times 3$ convolutions with deformable ones for high-level features (\textit{e.g.}, $conv_4$, $conv_5$) of ResNet-50/101 to model the transformation. But it does not show performance improvement for detection networks with group convolution or SE-block. The reason might be that the pre-trained features have been organized and the convolution filters in one group might not share the same deformable parameters. To tackle this, we introduce the independent aligned attention module. As illustrated in Fig.~\ref{fig_align_att}, a $1\times 1$ convolution layer is firstly used for dimensionality reduction. Then a $3 \times 3$ deformable convolution layer follows and another $1 \times 1$ convolution layer is added to equalize the dimension for residual unit. This newly-added aligned attention module can organize the convolution filters for shared deformable parameters learning. We add the aligned attention for feature map alignment in top-down feature merging. Besides, the aligned attention helps the learned networks focus more on foreground regions for better object representations.

\subsection{Hybrid attention and joint learning}
We further integrate the above attention modules together and jointly learn the hybrid attention. The feature maps from the backbone networks $C_x$ are firstly transformed with aligned attention, to generate $C'_x$. Then several convolution layers with CLGN-based channel attention are added to produce enhanced features based on the aligned attended features. Another branch with stacked dilated convolution is added to learn spatial attention, whose weights are further multiplied with the enhanced features. Finally, the channel attention with CLSE is added to act as soft feature selection before classification and regression. From the empirical study, we find that the non-linear mapping between channel and spatial attention helps improve the detection accuracy. Thus another convolution layers with $3\times 3$ kernel size with ReLU activation is added between spatial and the CLSE-based channel attention. With hybrid attention, we can improve the feature representations with focused learning. As noted before, the attention outputs soft weights to be multiplied with the feature responses from main network. The gradients are multiplied with the decayed coefficients, which might slow down the convergence of the whole network. To tackle with the slow learning, we firstly learn the main network with several iterations, then learn the attention network with another several iterations. The detection performance can be further improved with this learning.

\section{Network architecture for object detection}\label{sec:net}
To implement the hybrid attention mechanism, we choose FPN-Retina-Net as the baseline and integrate the attention modules for object detection. We further introduce the proposed network architecture for single-stage object detection.

\subsection{Retina-Net Baseline}
We use the single-stage FPN-Retina-Net~\cite{Lin_retina17} as the baseline, which is a fully-convolutional network with feature pyramid for object detection. For an image with fixed resolution, multi-level features are extracted based on the backbone network. Then a feature pyramid with top-down pathway and lateral connections is built. Each point in either level of feature response maps corresponds to several anchors with pre-defined locations. For Retina-Net, anchors are distributed in 2 scales with 3 kinds of aspect ratios as 1:1, 1:2, 2:1. For each level in the pyramid, the head net with two branches of four $3 \times3$ convolutional layers are attached. Another $3 \times 3$ convolutional layer follows after each branch for classification and bounding box regression, respectively. Focal loss is applied to tackle with imbalanced classification issue. Retina-Net is a highly-efficient single-stage detector without separate RPN and ROI-pooling. We choose Retina-Net as a strong baseline to further evaluate the performance of our hybrid attention.

\begin{figure}[htbp]
\centering
\includegraphics[width=3.2in]{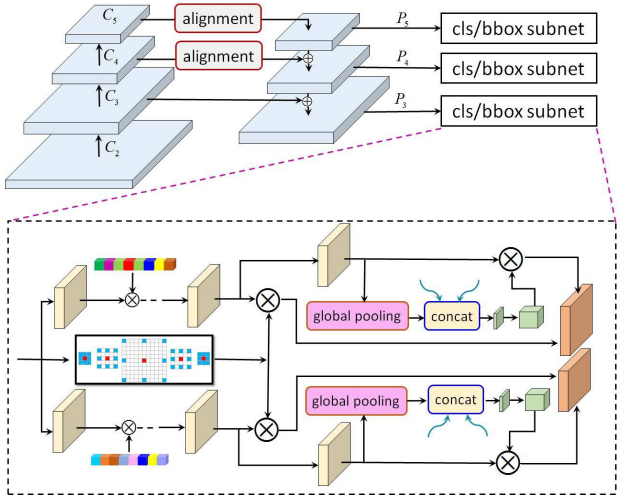}
\caption{Outline of the HAR-Net. Based on FPN-Retina-Net, we sequentially embed the aligned attention, channel attention and spatial attention for hybrid attention. The HAR-Net can utilize
multiple visual attention mechanism for effective single-stage object detection. {\label{fig_net}}}
\end{figure}

\subsection{HAR-Net: hybrid attention based Retina-Net}\label{sec:har-net}
As mentioned above, the hybrid attention can be easily embedded into CNNs. We plug the three kinds of attention modules into the single-stage Retina-Net for object detection. The modified network is named as HAR-Net, abbreviated for \textit{Hybrid Attention based Retina-Net}. The whole network is illustrated in Fig.~\ref{fig_net}. Based on the multi-level features $C_x$ from the backbone network, the aligned attention module follows afterwards to generate the aligned multi-level features as $C'_x$. In particular, we add the aligned attention module after $C_4$ and $C_5$ to align the high-level low-resolution features for merging. A feature pyramid is constructed with the aligned features $C'_x$. The outputs from each level in the pyramid are extracted as the feature responses for different scales $P_x$, with $2\times$ resolution changes. Then the head net with two branches of several convolution layers for classification and bounding box regression are connected after $P_x$, with sharing parameters across different levels. The channel and spatial attention is integrated into this head net. The CLGN with attached feature scaling is imposed after each convolution layer. In parallel with the two branches of stacked convolutional layers, another spatial attention branch with stacked dilated convolutions is added. The spatial attention weights are multiplied with each point in the response maps. In the end, an extra convolution layer with ReLU activation is added. The CLSE with concatenated global pooling outputs is connected afterwards to obtain the pyramid-shared channel attention for channel-wise multiplication.

In training phase, we use the images with annotated object boxes to learn the network parameters. It is noteworthy that the hybrid attention is embedded, therefore the HAR-Net can be learned end-to-end without additional annotations. Multi-task losses on classification/regression outputs are used to supervise both the attention and feature learning. However, the spatial and channel attention are both formulated as output weights less than 1 multiplying with feature maps. The learning of network parameters might be slowly converged. To tackle with the issue, another incremental training is introduced. That is, we firstly train the main network for better descriptive features, then add the spatial and channel attention modules with several more iterative learning. Both the end-to-end and incremental learning works well in practical, yet the latter can obtain minor improvement in accuracy.

\begin{figure}[htbp]
\centering
\includegraphics[width=3.0in]{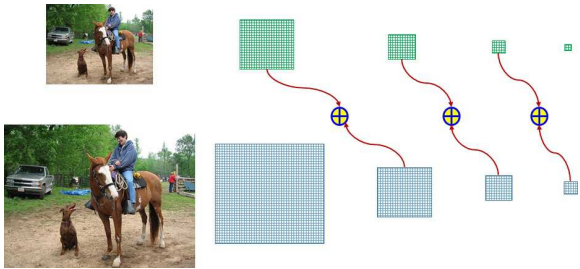}
\caption{Multi-scale testing with $2 \times$ resolution changes. If the anchors are spread over 5 scales, then the output maps of 4 scales are shared. We merge the output maps with the same resolutions for better prediction.
{\label{fig_dst}}}
\end{figure}

In testing phase, the output probability maps are used to inference the object existence. The anchors with confidence higher than a fixed threshold are stored as object candidates. Bounding box regression outputs are used to predict the object locations. Post-processing techniques are applied to remove repeated detections and refine the detection results, includes: \\
\textbf{Soft-NMS.} Instead of conventional non-maximal suppression (NMS), we apply Soft-NMS~\cite{Bodla_softnms17} to deal with the repeated detections. Different from conventional NMS which recursively eliminates the boxes with high IoU (intersection over union) and low confidence, Soft-NMS decays the confidence of all other detected boxes with a continuous function of the IoU. Since no detection is eliminated in this process, Soft-NMS can save the detections with high confidence and high IoU in crowd scenes.  \\
\textbf{Bounding box voting.} We apply the bounding box voting for precise localization. Suppose there are several overlapped detection boxes and the regressed bounding box $R_i$ is with probability $p_i$. Then the voted bounding box is denoted as
\begin{equation}
\begin{array}{c}
\widetilde{R}=\sum_i p_i R_i \left/\sum_i p_i\right. \\
\widetilde{p}=\max_i p_i
\end{array}
\label{eq3}
\end{equation} \\
Bounding box voting improves the localization accuracy by fusing multiple high-confidence detections. \\
\textbf{Multi-scale testing.} In single-stage detectors like Retina-Net, anchors are designed across several discrete scales. Like Retina-Net, HAR-Net is with anchors across $5$ scales, focusing on objects from $2^5$ to $2^{10}$ pixels. But it would be insufficient to detect object from varying scales. Motivated by this, we develop the multi-scale testing for single-stage detectors. With a trained network, we choose two scales with $2\times$ resolution changes and obtain the output maps with probabilities/box regression values at the two scales. As shown in Fig.~\ref{fig_dst}, the output maps from the two scales share the resolutions of $4$ scales. Thus we merge the output probabilities and regression values by average to predict the final detections. Besides, the remaining output maps from the smaller image are used to find large objects, while those from the larger image are used to find quite small objects. This double-scale testing is quite effective for detecting objects with scattered scales. It can easily be modified to multi-scale testing with $2 \times$ resolution changes. For multi-scale testing with random scales, we combine the detections for higher recall.

\section{Experimental results}\label{sec:exp}
In this section, we evaluate the attention modules and HAR-Net on \textit{COCO} object detection dataset~\cite{Lin_coco14}. It contains more than 100k images over 80 object categories. We follow the common experimental setting with around 115k images in the training set and the other 5k images in the validation set for algorithm study and meta-parameter setting. Throughout all experiments, we compare with the state-of-the-art methods on the dataset without bells and whistles.

Our programs are implemented by Caffe Toolkit~\cite{Jia_caffe14}. Unless otherwise specified, the networks are trained with 10 NVidia Titan-X GPUs, by synchronized SGD with 2 images per GPU. The network is trained for 48k iterations with learning rate 0.0125. Then the learning rate is divided by 10 for another 16k iterations and again for the last 8k iterations. We set the weight decay to 0.0001 and the momentum to 0.9. We firstly train the Retina-Net to obtain the baseline accuracy. The attention modules are added progressively for separate evaluation. We mainly employ the ImageNet-pretrained ResNet-50/101~\cite{He_resnet16} as the backbone networks for feature representations. We also add the experiments over the ImageNet-pretrained SENets~\cite{Hu_senet18} for competitive performance. The mean Average Precision (mAP) over different IOUs across 80 categories is used as the measurements. Without noticing, we use single-scale training and testing. Scale jittering is applied and horizontal image flipping is used for data augmentation. Soft-NMS~\cite{Bodla_softnms17} and bounding box voting are set as the common post-processing steps.

\subsection{Spatial attention}

We firstly investigate the effects of the spatial attention module with stacked dilated convolutions. Retina-Net with ResNet-50 as the backbone network is constructed. The spatial attention module with different dilation settings is added and the \textit{mAP} to measure the detection accuracy is listed in Table~\ref{table_sa} for comparison. With the baseline Retina-Net, the detection accuracy is 35.1\% of $AP_{50:90}$ and 52.9\% of $AP_{50}$. From the table we can see that the spatial attention learning is quite effective to improve the single-stage detection accuracy. We set $C=32$ in spatial attention branch. Even with three common $3 \times 3$ convolution layers for spatial attention learning, the achieved $AP_{50:90}$ is 36.4\%, improving by 1.3 points. We further adjust the dilation rates of stacked convolution layers. When the dilation rates change in an increase way as \textit{1-2-5}, the achieved $AP_{50:90}$ is still 36.4\%. When the dilation rates decrease as \textit{5-2-1}, the $AP_{50:90}$ and $AP_{50}$ is 36.5\% and 54.6\%, which is higher. It is mainly because the reduced dilation setting for spatial attention learning helps alleviate the gridding artifacts. Furthermore, the dilation rates of spatial attention module are set to be symmetrical, as \textit{1-2-5-2-1}. For fair comparison, we reduce the convolution channels number to $C=16$ to keep the parameter size. The detection accuracy is slightly higher, with $AP_{50:90}$ as 36.6\% and $AP_{50}$ as 54.9\%. It is noteworthy that the spatial attention mainly improves the detection accuracy of small objects, but the accuracy of large objects slightly increases. Compared to baseline, the $AP_S$ improves from 16.4\% to 18.9\%, by 2.5 points. It is reasonable since the spatial attention mainly learns the relative importance of regions in scenes with larger receptive fields. It also helps bridge the feature gap of small objects by adding context information.

\begin{table}[htbp]
\begin{center}
\caption{Comparison for stacked dilated convolution based spatial attention (SA) with different dilation settings. The shorter side of input images is normalized to 600 pixels, with longer side less than 1000 pixels. The backbone network is ResNet-50.} \label{table_sa}
\begin{small}
\begin{tabular}{p{20mm}|p{9mm}|p{6mm}|p{6mm}|p{6mm}|p{6mm}|p{6mm}}
\hline
Methods &$AP_{50:95}$ &$AP_{50}$ &$AP_{75}$ &$AP_S$	&$AP_M$ &$AP_L$ \\ \hline
Baseline &35.1 &52.9 &37.9 &16.4 &39.0 &50.7 \\
\textit{SA(1-1-1)} &36.4 &54.4 &39.7 &18.5 &40.7 &50.9  \\
\textit{SA(1-2-5)} &36.4 &54.4 &39.6 &18.4 &40.8 &51.0 \\
\textit{SA(5-2-1)} &36.5 &54.6 &\textbf{39.8} &\textbf{18.9} &\textbf{40.9} &\textbf{51.8} \\
\textit{SA(1-2-5-2-1)} &\textbf{36.6} &\textbf{54.9} &\textbf{39.8} &18.6 &\textbf{40.9} &51.3 \\ \hline
\end{tabular}
\end{small}
\end{center}
\end{table}

\subsection{Channel attention}

\begin{table}[htbp]
\begin{center}
\caption{Comparison for different kinds of channel attention (CA), with ResNet-50 as the backbone. The shorter side of test images is normalized to 600 pixels, with longer side less than 1000 pixels. \textit{SE-CA} indicates separate SE-block for channel attention in pyramid levels. \textit{CLSE-CA} indicates pyramid-sharing channel attention with CLSE. \textit{CLSEGN-CA} indicates pyramid-sharing channel attention with CLSE and CLGN.}\label{table_fa}
\begin{small}
\begin{tabular}{p{20mm}|p{9mm}|p{6mm}|p{6mm}|p{6mm}|p{6mm}|p{6mm}}
\hline
Methods &$AP_{50:95}$ &$AP_{50}$ &$AP_{75}$ &$AP_S$	&$AP_M$ &$AP_L$ \\ \hline
Baseline &35.1 &52.9 &37.9 &16.4 &39.0 &50.7 \\
\textit{SE-CA} &36.4 &54.4 &39.7 &18.4 &40.6 &51.7 \\
\textit{CLSE-CA} &36.6 &54.7 &39.8 &\textbf{18.8} &41.1 &51.4 \\
\textit{CLSEGN-CA} &\textbf{37.1} &\textbf{55.6} &\textbf{40.4} &\textbf{18.8} &\textbf{41.6} &\textbf{52.0} \\ \hline
\end{tabular}
\end{small}
\end{center}
\end{table}

We evaluate different ways of channel attention learning and the detection accuracy is presented in Table~\ref{table_fa}. The first kind \textit{SE-CA} is to add separate SE-blocks after the last convolution layers, with a single SE-block attached with either branch at every level in the pyramid. It indicates that the channel attention is learned independently for different levels in the pyramid. Compared with the baseline, \textit{SE-CA} increases the $AP_{50:90}$ from 35.1\% to 36.4\%, and $AP_{50}$ from 52.9\% to 54.4\%. It shows that channel attention is useful for single-stage object detection. Since the parameters of convolution are shared across multiple levels, we further use the concatenation of global pooling outputs from all levels in the pyramid to learn the cross-level channel attention, denoted as \textit{CLSE-CA}. This pyramid-sharing of channel attention learning further improves the detection accuracy. The \textit{CLSE-CA} achieved $AP_{50:90}$ is 36.6\% and $AP_{50}$ is 54.7\%. We finally add the CLGN and attached pyramid-sharing feature scaling for channel attention of convolutional filters, denoted as \textit{CLSEGN-CA}. This is exactly the same as the channel attention module described in subsection~\ref{sec:fe_att}. From the table we can find that the detection accuracy is further increased. The $AP_{50:90}$ and $AP_{50}$ of \textit{CLSEGN-CA} is 37.1\% and 40.4\%, respectively. Compared to the baseline, channel attention increases the $AP_{50:90}$ and $AP_{50}$ relatively by 5.7\% and 5.1\%, respectively.

\subsection{Aligned attention}
We also test the aligned attention for the feature pyramid construction. The aligned attention modules are progressively added into different levels in the pyramid and the detection accuracy is presented in Table~\ref{table_aa} for comparison. Since $C_5$ impacts all levels of features in the pyramid, we firstly attach $C_5$ with aligned attention. The achieved $AP_{50:95}$ is 36.7\% and $AP_{50}$ is 54.8\%. It is reasonable that the high-level features in pyramid have larger receptive fields but low resolution. The aligned attention increases the resolutions of high-level feature maps with deformation, therefore improves the pyramid feature representations for object detection. Then the aligned attention is added for both $C_4$ and $C_5$. The $AP_{50:95}$ is 36.9\% and $AP_{50}$ is 55.0\%. Compared to the baseline, the $AP_{50:95}$ is increased by 1.8 points and $AP_{50}$ by 2.1 points. We also add the aligned attention after all levels of features from the backbone networks, as $C_3, C_4, C_5$. But the detection accuracy is lower, with $AP_{50:95}$ of 36.8\%. It might be that the alignment of $C_3$ is less important for detection, meanwhile the learning would be insufficient in pyramid streams. Based on this empirical study, we add aligned attention for $C_4$ and $C_5$, to balance the detection accuracy and computational efficiency.

\begin{table}[htbp]
\begin{center}
\caption{Comparison of aligned attention (AA) for different levels of features, with ResNet-50 as the backbone. The shorter side of images is normalized as 600 pixels, with longer side less than 1000 pixels.} \label{table_aa}
\begin{small}
\begin{tabular}{p{20mm}|p{9mm}|p{6mm}|p{6mm}|p{6mm}|p{6mm}|p{6mm}}
\hline
Methods &$AP_{50:95}$ &$AP_{50}$ &$AP_{75}$ &$AP_S$	&$AP_M$ &$AP_L$ \\ \hline
Baseline &35.1 &52.9 &37.9 &16.4 &39.0 &50.7 \\
\textit{AA}($C_5$) &36.7 &54.8 &39.8 &19.2 &40.9 &51.6 \\
\textit{AA}($C_4,C_5$) &\textbf{36.9} &\textbf{55.0} &\textbf{40.1} &\textbf{19.3} &\textbf{41.3} &52.2 \\
\textit{AA}($C_3,C_4,C_5$) &36.8 &54.9 &39.8 &18.8 &40.8 &\textbf{52.4} \\ \hline
\end{tabular}
\end{small}
\end{center}
\end{table}

\subsection{Hybrid attention}

Finally, we present the ablation study of hybrid attention and HAR-Net with different backbone networks in Table~\ref{table_ha}. In this experiment, the shorter side of test images is set to 800 pixels with the longer side less than 1200 pixels for training and single-scale testing. For multi-scale testing, we set the short sides as $\left\{500,750,1500 \right\}$, where the latter two scales are with $2\times$ changes. The detection accuracy is placed in the order as ¡°\textit{baseline}$\rightarrow$\textit{AA+CA}$\rightarrow$\textit{HA}$\rightarrow$\textit{HA w ms}¡±. From the table we can find that the joint learning of hybrid attention is effective for single-stage object detection. With ResNet-50 as the backbone network, the baseline Retina-Net achieves $AP_{50:95}$ of 36.0\%. The detection accuracy is increased to 38.6\% with aligned attention and channel attention. The final ResNet-50-based HAR-Net achieves the $AP_{50:95}$ of 38.9\% with single-scale testing and 40.9\% with multi-scale testing. Compared to the baseline, the detection accuracy is increased by 4.9 points. With ResNet-101 as the backbone network, the HAR-Net achieves the single-scale detection $AP_{50:95}$ of 40.6\% and multi-scale detection $AP_{50:95}$ of 43.1\%. The detection accuracy is increased by 5.1 points compared to the baseline Retina-Net. Besides, we also evaluate the SE-ResNet-based HAR-Net. The detection AP is increased by 2.0$\sim$2.8 points with hybrid attention and another 1.9$\sim$2.2 points with multi-scale testing. In particular, with SE-ResNeXt-101 as the backbone network, the achieved $AP_{50:95}$ is 42.7\% of single-scale testing and 45.3\% with multi-scale testing, which is the state-of-the-art single-stage object detection accuracy on \textit{COCO 5k-validation} dataset.

\begin{table*}[htbp]
\begin{center}
\caption{Ablation study of hybrid attention over different backbone networks. The $AP$ is placed in ¡°baseline$\rightarrow$AA+CA$\rightarrow$HA$\rightarrow$HA \textit{w ms}¡± order.} \label{table_ha}
\begin{small}
\begin{tabular}{p{25mm}|p{40mm}|p{40mm}|p{40mm}}
\hline
Backbone &$AP_{50:95}$ &$AP_{50}$ &$AP_{75}$ \\ \hline	
ResNet-50 &36.0$\rightarrow$38.6$\rightarrow$38.9$\rightarrow$\textbf{40.9}
&54.2$\rightarrow$57.1$\rightarrow$57.9$\rightarrow$\textbf{60.1}
&39.1$\rightarrow$42.0$\rightarrow$42.7$\rightarrow$\textbf{45.1} \\
ResNet-101 &38.0$\rightarrow$40.1$\rightarrow$40.6$\rightarrow$\textbf{43.1}
&56.5$\rightarrow$58.9$\rightarrow$60.3$\rightarrow$\textbf{62.9}
&41.4$\rightarrow$43.9$\rightarrow$45.0$\rightarrow$\textbf{47.5} \\
SE-ResNet-50 &36.8$\rightarrow$39.5$\rightarrow$39.5$\rightarrow$\textbf{41.7}
         &57.5$\rightarrow$59.0$\rightarrow$58.9$\rightarrow$\textbf{61.4}
        &39.9$\rightarrow$42.9$\rightarrow$43.4$\rightarrow$\textbf{45.8}\\
SE-ResNet-101 &39.1$\rightarrow$41.6$\rightarrow$41.9$\rightarrow$\textbf{43.8}
&58.8$\rightarrow$61.1$\rightarrow$61.2$\rightarrow$\textbf{63.6}
&42.8$\rightarrow$45.3$\rightarrow$45.9$\rightarrow$\textbf{48.0} \\
SE-ResNeXt-50 &38.7$\rightarrow$40.6$\rightarrow$40.8$\rightarrow$\textbf{42.7}
&58.7$\rightarrow$59.8$\rightarrow$60.2$\rightarrow$\textbf{62.5}
&42.6$\rightarrow$44.1$\rightarrow$45.0$\rightarrow$\textbf{46.9}  \\
SE-ResNeXt-101 &41.2$\rightarrow$42.7$\rightarrow$43.2$\rightarrow$\textbf{45.3}
&60.7$\rightarrow$62.1$\rightarrow$62.7$\rightarrow$\textbf{64.7}
 &45.1$\rightarrow$46.5$\rightarrow$47.1$\rightarrow$\textbf{49.6} \\ \hline
\end{tabular}
\end{small}
\end{center}
\end{table*}

\begin{table*}[htbp]
\begin{center}
\caption{Comparison of HAR-Net with the state-of-the-art object detection methods on \textit{COCO-test-dev} dataset. The \underline{AP}s highlights the highest single-scale accuracy and the \textbf{AP}s highlight the highest multi-scale accuracy. } \label{table_coco_test}
\begin{small}
\begin{tabular}{p{36mm}|p{35mm}|p{9mm}|p{8mm}|p{8mm}|p{7mm}|p{7mm}|p{7mm}}
\hline
Methods &backbobe &$AP_{50:95}$ &$AP_{50}$ &$AP_{75}$ &$AP_S$ &$AP_M$ &$AP_L$ \\ \hline
\textit{Multi-stage methods}&\multicolumn{7}{c}{} \\  \hline
Faster-RCNN~\cite{Ren_fasterrcnn15}&ResNet-101 &36.7 &54.8 &39.8 &19.2 &40.9 &51.6 \\
R-FCN~\cite{Dai_rfcn16} &ResNet-101 & 35.1 &52.9 &37.9 &16.4 &39.0 &50.7 \\
Deformable R-FCN~\cite{Dai_deform17}& Aligned-Inception-ResNet &37.5 &58.0 &40.8 &19.4 &40.1 &52.5 \\
Faster RCNN w FPN~\cite{Lin_FPN17} &ResNet-101 &36.2 &59.1 &39.0	 &18.2 &39.0 &48.2 \\
Mask RCNN~\cite{He_maskrcnn17} &ResNet-101 &38.2 &60.3 &41.7 &20.1 &41.1 &50.2 \\
Mask RCNN~\cite{He_maskrcnn17} &ResNeXt-101 &39.8 &62.3 &43.4 &22.1 &43.2 &51.2 \\
TDM~\cite{Shrivastava_tdm16} &Inception-ResNet-v2	&36.8 &57.7 &39.2 &16.2 &39.8 &52.1 \\
AttractioNet~\cite{Gidaris_attractionNet16} &VGG16+Wide ResNet &35.7 &53.4 &39.3 &15.6 &38.0 &52.7 \\
Cascade RCNN~\cite{Cai_cascadeRCNN18} &ResNet-101 &42.8 &62.1 &46.3 &23.7 &45.5 &55.2 \\ \hline
\textit{Single-stage methods}&\multicolumn{7}{c}{} \\  \hline
YOLO-v3~\cite{Redmon_yolov3_18} &DarkNet-53	 &33.0 &57.9 &34.4 &18.3 &35.4 &41.9 \\
SSD513~\cite{Liu_ssd14} &ResNet-101 &31.2	&50.4 &33.3 &10.2 &34.5 &49.8 \\
DSSD513~\cite{Fu_DSSD17}	 &ResNet-101 &33.2 &53.3 &35.2 &13.0 &35.4 &51.1 \\
Retina-Net~\cite{Lin_retina17} &ResNet-101	 &39.1 &59.1	&42.3 &21.8	&42.7 &50.2 \\
Retina-Net~\cite{Lin_retina17} &ResNeXt-101 &40.8	&61.1 &44.1 &24.1 &44.2 &51.2 \\
GHM-Retina-Net~\cite{Li_ghm19} &ResNet-101 &39.9 &60.8 &42.5 &20.3 &43.6	&54.1 \\
GHM-Retina-Net~\cite{Li_ghm19} &ResNeXt-101 &41.6	&62.8 &44.2 &22.3	 &45.1 &55.3 \\
FSAF~\cite{Zhu_fsaf19} &ResNet-101 &40.9 &61.5 &44.0 &24.0 &44.2	&51.3 \\
FSAF \textit{w ms}~\cite{Zhu_fsaf19} &ResNet-101 &42.8 &63.1 &46.5 &27.8 &45.5 &53.2 \\
FSAF~\cite{Zhu_fsaf19} &ResNeXt-101 &42.9 &63.8 &46.3 &\underline{26.6} &46.2 &52.7 \\
FSAF \textit{w ms}~\cite{Zhu_fsaf19} &ResNeXt-101 &44.6 &65.2 &48.6 &\textbf{29.7} &47.1 &54.6 \\
HAR-Net &ResNet-101 &41.1 &60.5 &45.0 &23.0 &44.8 &52.8 \\
HAR-Net \textit{w ms} & ResNet-101 &43.2 &63.0 &47.5 &26.0 &45.8 &53.9 \\
HAR-Net &SE-ResNeXt-101 &\underline{43.8} &\underline{63.4} &\underline{47.8} &24.9 &\underline{47.7} &\underline{56.4} \\
HAR-Net \textit{w ms} & SE-ResNeXt-101 &\textbf{45.8} &\textbf{65.8} &\textbf{50.4} &28.4 &\textbf{48.6} &\textbf{57.4} \\ \hline
\end{tabular}
\end{small}
\end{center}
\end{table*}

We further perform extensive comparison over HAR-Net with existing object detection methods on \textit{COCO test-dev} set. The object detection methods are divided into \textit{multi-stage methods} and \textit{single-stage methods}. The \textit{two-stage methods} like R-FCN~\cite{Dai_rfcn16} and Faster-RCNN~\cite{Ren_fasterrcnn15} are included in \textit{multi-stage methods}. We mainly compare the performance of HAR-Net with \textit{single-stage methods} without RPN and ROI-pooling. With ResNet-101 as the backbone network, the HAR-Net achieves the single-scale testing $AP_{50:95}$ of 41.1\% and multi-scale testing $AP_{50:95}$ of 43.2\%. Compared to ResNet-101-based GHM Retina-Net~\cite{Li_ghm19}, the $AP_{50:95}$ is increased by 1.2$\sim$3.3 points. The detection accuracy is also comparable with the \textit{multi-stage} cascade-RCNN~\cite{Cai_cascadeRCNN18}. With SE-ResNeXt-101 as the backbone network, the HAR-Net achieves $AP_{50:95}$ of 43.8\%, which is 7.4\%, 5.3\% and 2.1\% relatively higher than Retina-Net~\cite{Lin_retina17}, GHM-Retina-Net~\cite{Li_ghm19} and FASF~\cite{Zhu_fsaf19}, respectively. With multi-scale testing, $AP_{50:95}$ of SE-ResNeXt-101 based HAR-Net is 45.8\%. It proves that the proposed multi-scale testing is quite effective for \textit{single-stage detectors}. Compared to \textit{single-stage methods} such as YOLO~\cite{Redmon_yolov3_18} and SSD~\cite{Liu_ssd14}, the detection accuracy is improved with a large margin. Compared with the AttractioNet~\cite{Gidaris_attractionNet16}, the mAP is increased by around 10 points, which shows the effectiveness of our hybrid attention. Besides, the detection performance is also comparable with the object detection methods with scale normalization such as SNIP~\cite{Singh_snip18} and SNIPER~\cite{Singh_sniper18}. Note that the performance gain of scale normalization training is complementary to the network design and the performance will be further improved by such training techniques.

\section{Conclusion}\label{sec:con}
In this paper, we exploit the visual attention mechanism for single-stage object detection, especially for FPN-based detection. Three kinds of attention modules are proposed and sequentially integrated into object detection network. Spatial attention is learned with symmetrically dilated convolutions for soft region proposal, to alleviate the imbalanced classification. Channel attention is learned with CLGN and CLSE, for effective feature selection. Aligned attention is learned with an independent deformation module, for feature alignment in pyramid construction. We combine the three attention modules and propose the hybrid attention mechanism for single-stage object detection. The HAR-Net is proposed by integrating the hybrid attention into Retina-Net. Besides, we also develop the multi-scale testing for single-stage detectors. Experiments demonstrate the effectiveness of hybrid attention and the HAR-Net. In the future, we would like to combine the efficient multi-scale training techniques and develop the adaptive scale selection to further improve the single-stage object detection performance.


%




\ifCLASSOPTIONcaptionsoff
  \newpage
\fi



\bibliographystyle{IEEEtran}
\bibliography{egbib}
%

%


\begin{IEEEbiographynophoto}{Ya-Li Li}
(M'17)received the B.E. degree with Excellent Graduates Award from Nanjing University, China, in 2007 and the Ph.D degree from Tsinghua University, Beijing, China, 2013. Currently she is a research assistant in Department of Electronic Engineering, Tsinghua University. Her research interests include image processing, pattern recognition, computer vision, and video analysis etc.
\end{IEEEbiographynophoto}


\begin{IEEEbiographynophoto}{Shengjin Wang}
(M'04) received the B.E. degree from Tsinghua University, China, in 1985 and the Ph.D. degree from the Tokyo Institute of Technology, Tokyo, Japan, in 1997. From May 1997 to August 2003, he was a member of the research staff in the Internet System Research Laboratories, NEC Corporation, Japan. Since September 2003, he has been a Professor with the Department of Electronic Engineering, Tsinghua University, where he is currently also a Director of the Research Institute of Image and Graphics (RIIG). He has published more than 50 papers on image processing. He is the holder of ten patents. His current research interests include image processing, computer vision, video surveillance, and pattern recognition.
\end{IEEEbiographynophoto}




\end{document}